\documentclass[10pt,twocolumn,letterpaper]{article}

\usepackage{cvpr}
\usepackage{times}
\usepackage{epsfig}
\usepackage{graphicx}
\usepackage{amsmath}
\usepackage{amssymb}
\usepackage{booktabs}
\usepackage{authblk}

\usepackage[breaklinks=true,bookmarks=false]{hyperref}

\cvprfinalcopy % *** Uncomment this line for the final submission

\def\cvprPaperID{****} % *** Enter the CVPR Paper ID here

\begin{document}

%%%%%%%%% TITLE
\title{Adaptive Name Entity Recognition  under Highly Unbalanced Data}

% \author{Thong Thac Nguyen\\
% \textit{Saarland University }\\
% \and
% Duy Ho Minh Nguyen\\
% \textit{Saarland University}\\
% \and
% Pramod Rao \\
% \textit{Saarland University}\\
% }

\author[*1]{Thong Nguyen}
\author[*1]{Duy Nguyen}
\author[1]{Pramod Rao \thanks{This work was done as a project for the course ``Neural Networks Theory and Implementation" at Saarland University during the Winter Semester 2019/20.
% All the authors have contributed equally.
}}
\affil[1]{Saarland University, Saarland Informatics Campus, Germany \authorcr Email: {\tt \{s8tgnguy, s8hhnguy, s8prraoo\}@stud.uni-saarland.de}\vspace{1.5ex}}
\maketitle
%\thispagestyle{empty}

%%%%%%%%% ABSTRACT
\begin{abstract}
For several purposes in Natural Language Processing (NLP), such as Information Extraction, Sentiment Analysis or Chatbot, Named Entity Recognition (NER) holds an important role as it helps to determine and categorize entities in text into predefined groups such as the names of persons, locations, quantities, organizations or percentages, etc. In this report, we present our experiments on a neural architecture composed of a Conditional Random Field (CRF) layer stacked on top of a Bi-directional LSTM (BI-LSTM) layer for solving NER tasks. Besides, we also employ a fusion input of embedding vectors (Glove, BERT), which are pre-trained on the huge corpus to boost the generalization capacity of the model. Unfortunately, due to the heavy unbalanced distribution cross-training data, both approaches just attained a bad performance on less training samples classes. To overcome this challenge, we introduce an add-on classification model to split sentences into two different sets: Weak and Strong classes and then designing a couple of Bi-LSTM-CRF models properly to optimize performance on each set. We evaluated our models on the test set and discovered that our method can improve performance for Weak classes significantly by using a very small data set (approximately 0.45\%) compared to the rest classes.
\end{abstract}

%%%%%%%%% BODY TEXT
\section{Introduction}

Named Entity Recognition (NER), which represents the task of detecting the boundary and type of named entities, plays a crucial role in many applications in the field of natural language processing. In conversational question answering, answering factoid questions like \textit{``Who is the president of United States?''} requires the interested entity, such as \textit{``United States''}, to be detected and linked to corresponding nodes in knowledge graph where the answer is stored. Detecting and linking named entities also help to enrich short queries which are highly ambiguous and lack of clarity, which in turn improving the performance of search engine systems. 

Depending on specific applications, the set of entity types might vary. However, the label set most commonly used in academic research contains four labels, including LOC(location), ORG(organization), PER(person) and MISC(miscellaneous). In this project, the given corpus was annotated with $8$ types of entities: Art(artifact), Eve(event), Nat(natural phenomena), Geo(geographical entity), Gpe(geopolitical entity), Tim(time indicator), Org(organization), Per(person). The statistic of number of annotations for each entity is reported in Table \ref{tab:lab_stat}.

As many other language processing tasks, the first challenge of NER task is the high ambiguity of human written languages. Entities with the same text surface might belong to different entity types. In sentence, ``\textit{The US president, Barack Obama invited Shinzō Abe to visit US next April}'', the former \textit{``US''} is a geopolitical entity, while the later \textit{``US''} is a geographical entity. Further more, the complexity of NER system increases as the size of label set increases. Larger label set comes along with more uncertainty and prediction errors. To add to that, the entity set usually varies from domain to domain, therefore it is extremely hard to build a general named entity tagger, and building domain specific system requires a lot of effort in creating new annotated training data. 

Previous work on NER task could be grouped into two main approaches which are traditional feature-based approach and neural network approach. The former one utilizes traditional machine learning models, such as Conditional Random Fields (CRF) \cite{CRF}, Support Vector Machine (SVM) \cite{SVM}, Maximum Entropy (ME)\cite{ME}, which needs related features designed by expert. With the increasing availability of data and computing resources, the later approach has emerged and become very successful recent years. The neural network approach mitigates the concern of tedious and labour-intensive feature engineering work of the traditional approach, but requires a large amount of annotated data and also need architecture design carried out by specialized experts. The common architectures include RNN \cite{Graves}, (Bi-)LSTM \cite{Alex}, (Bi-)LSTM-CRF \cite{Huang}, Transformer \cite{Vaswani}, BERT \cite{Devlin}, etc. 

In this project, we implemented Bi-LSTM-CRF architecture using Pytorch framework, and we also experimented with one traditional model using CRF (Stanford NLP) and one SOTA transformer based neural network model (BERT). The best result among above models was 0.85 F1 score on test set. However, these models all suffered from the problem of imperfect annotations (wrong annotations) and highly imbalanced dataset. The Figure \ref{fig:Distribution} and Table \ref{tab:lab_stat} depict the distribution for all 9 classes (including class O(other)) reported from the whole train/test/val corpus. It could be seen that the distribution of labels is highly skewed. Specifically, the total samples of five groups (Strong class): ``Geo'', ``Time'', ``Org'', ``Per'' and ``Gpe''  is approximately 50 times more than the total samples of the last three groups (Weak class): ``Art'', ``Eve'' and ``Nat''. This characteristic poses a considerable difficulty for almost every models as most of the features learned are biased to the Strong classes. Therefore, in latter part of this report, we will also describe an approach to solve this problem.
 \begin{table}[h!]
 \begin{center}
          \begin{tabular}{|c|c|c|c|}
     \hline
     \textbf{Label} & \textbf{Count} & \textbf{Label} & \textbf{Count} \\
     \hline
     \hline 
         O &  88791 & Geo & 37644 \\
         Tim & 20333 & Org & 20143 \\
         Per & 16990 & Gpe & 15869\\
         Art & 402 & Eve & 308 \\
         Nat & 201 & &  \\
    \hline
     \end{tabular}
     \label{tab:lab_stat}
 \end{center}
          \caption{Label statistic}

 \end{table}

The rest of this report is organized as following:
\begin{itemize}
    \item Section 2 describes the overall methodology and several neural network architectures used in this work. In particular, Section 2.1 includes a very simple feed-forward neural network NER tagger which is used as a baseline model and also the Bi-LSTM-CRF models. Next in Section 2.2, we present how can we transfer learning from BERT model with a linear layer for the NER tag prediction. The Section 2.3 introduces our approach to detect the Weak class in a sentence based on RNN-CNN network. Finally, Section 2.4 briefly describes majority voting technique that is used to make the final prediction. 
    \item In Section 3, we summarize the main experiment results on both binary classification and NER task with baseline method discussed in section 2. 
    \item Finally, we devote the Section 4 and 5 to discuss the main contributions, discuss the future work and give the conclusion.
\end{itemize}

%------------------------------------------------------------------------
\section{Method}
\begin{figure}[t]
	\begin{center}
		\includegraphics[width=0.4\textwidth]{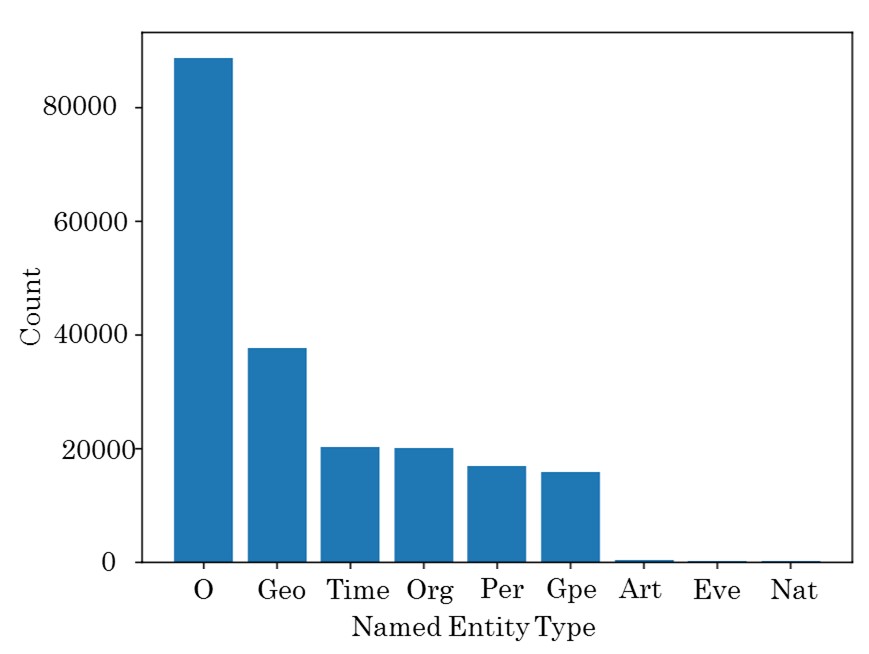}
	\end{center}
	\caption{Data distribution over all entity types.}  
\label{fig:Distribution}
\end{figure}

%Describe a main figure for illustrating our flow chart. 
We propose a new methodology for handling unbalanced classes in NER tagging task. Our method utilizes Bi-LSTM-CRF network for NER tagging. See Figure \ref{fig:Distribution} for an overview of the proposed framework. The proposed architecture uses an embedding layer that extracts word-level embedding feature vector from the sentences and is given as input to the Bi-LSTM-CRF network, this network predicts the NER tags for the Strong classes. On top of that, we use RNN-CNN network to detect tags from Weak class (``Art'', ``Eve'' and ``Nat''). This network is trained using Glove embedding pre-trained from several huge corpus of text. After identifying the Weak tags, only those sentences are picked and given to the Bi-LSTM-CRF network to obtain predictions on Weak classes. Finally, the predictions from the Strong class and Weak class are combined my majority voting method. In addition to Bi-LSTM-CRF model, we experiment with BERT model with a feed-forward layer as the output. The pre-trained BERT model with a feed-forward layer is used instead of Bi-LSTM-CRF network without any other changes to the existing framework. In the following sections we describe each module in greater detail.
\begin{figure*}[h]
	\begin{center}
		\includegraphics[width=0.9\textwidth]{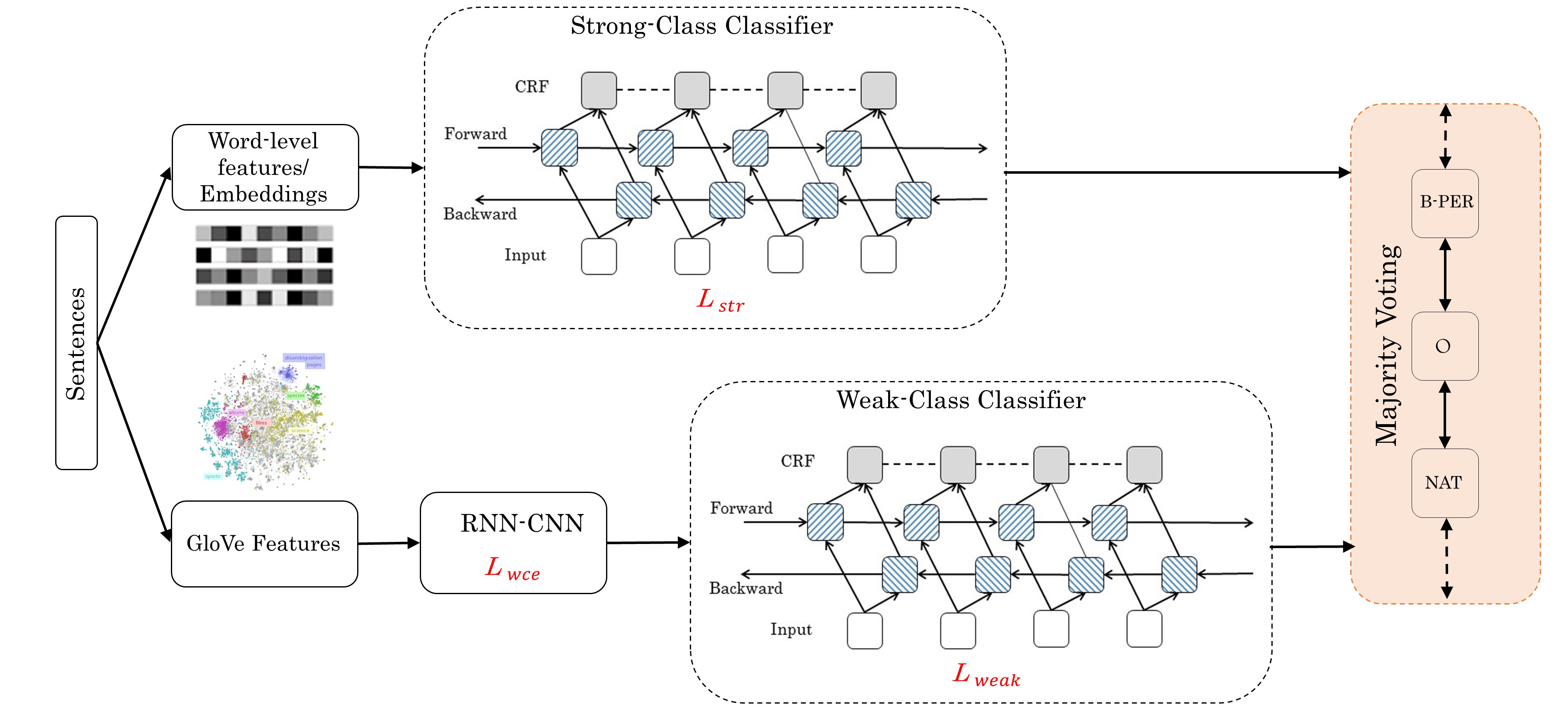}
	\end{center}
	\caption{Overview of the proposed method for NER tagging task under highly imbalanced data. In order to predict the Strong class tags Bi-LSTM-CRF is used, \textbf{$L_str$} represents the loss for the Strong class. Also, for the Weak class prediction, we use RNN-CNN network with a weighted cross entropy loss \textbf{$L_wce$} as a Weak class detector followed by Bi-LSTM-CRF model to predict Weak class tags. \textbf{$L_str$} represents the loss for the Weak class. These two predictions are finally combined by majority voting technique to get the final prediction}
\end{figure*}
%------------------------------------------------------------------------

\subsection{Label Sequence based on Bi-LSTM and CRF}
\label{subsection: BI-LST}
Initial experiments with simple a feed-forward network with three layers and ReLU \cite{ReLU} activation function in conjunction with the embedding layers for encoding inputs did not yield good results. To encourage exploitation of sequence data and share parameters we use sequence neural networks.

In modern-day datasets that frequently have lengthy sentences, vanilla recurrent neural networks might suffer from vanishing or exploding gradients \cite{Sepp} issues that severely handicap the learning process. This prompted us to use bidirectional LSTM (Bi-LSTM) that can model the long-term dependencies to capture both future and past context of the word.

Finally, to predict the NER tag for a word, we use a discriminative method of classifers, Conditional Random Fields (CRFs), that model the conditional distribution $p(y|x)$. CRFs focuses on the sentence level instead of individual positions for predicting a multivariate output $\textbf{y} = \{y_1, y_2, ... y_T\}$ with a large number of input sentence, $\textbf{X} = \{x_1, x_2, ... x_n\}$. In this work, we use linear chain CRFs where $p(y|x)$ is given as:
\begin{equation}
    p_\theta(y|x) = \frac{1}{Z_\theta(x)}exp \left( \sum_{t=1}^T \sum_{k=1}^K \theta_k f_k(y_{t-1}, y_t, x_t)\right)
\end{equation}
where $f_k(y_t-1, y_t, x_t)$ is the function for the properties of transition from the state $y_{t-1}$ to $y_t$ with input $x_t$ and $\theta_k$ is the parameter that is optimized during training. $Z_\theta$ is the normalisation factor that is given by:
\begin{equation}
    {Z_\theta(x)} = \sum_{y} exp\left( \sum_{t=1}^T \sum_{k=1}^K \theta_k f_k(y_{t-1}, y_t, x_t)\right)
\end{equation}

The Bi-LSTM output produces a matrix $\textbf{P}$ of scores for a given input sequence. The CRF learns the transition probability of the output labels, $\textbf{A} \in \mathbb{R}^{k+2, k+2}$. K is the number of distinct labels, and +2 indicates one tag each for start and end marker. \textbf{$P_{ij}$} corresponds to $j^{th}$ tag of the $i^{th}$ word. For a sequence of predictions, we define its score to be:
\begin{equation}
S(\textbf{X},\textbf{y}) = \sum_{i=0}^T A_{y_i,y_{i+1}} + \sum_{i=1}^T P_{ij}
\end{equation}

The probability for the sequence y is given by:
\begin{equation}
    p(\textbf{y}|\textbf{X}) = \frac{exp(S(\textbf{X},\textbf{y})}{\sum_{y' \in \textbf{}} exp(S(\textbf{X},\textbf{y}'))}
\end{equation}
The objective function is the maximum likelihood
of the probability distribution denoted as:
\begin{equation}
    \ln p(\textbf{y}|\textbf{X}) = S(\textbf{X},\textbf{y}) - \ln \sum_{y' \in \textbf{y}} exp(S(\textbf{X},\textbf{y}'))
\end{equation}
The maximum likelihood of the probability of predictions are maximized during training and the final tag $y^*$ is given by:
\begin{equation}
    y^* = \arg \max_{y' \in \textbf{y}} S(\textbf{X}, \textbf{y})   
\end{equation}
It has been shown that combing CRFs with Bi-LSTM for modelling the tagging task improves the tagging accuracy in general by effectively incorporating several dependencies across the output labels. Our experiment results as shown in the Table \ref{tab:final_acc} are concurrent with existing literature.

% \textcolor{red}{Incl false positives and separate models}
Interestingly, it can be inferred from the same Table \ref{tab:final_acc} that prediction accuracy for the Weak classes is poor compared to the Strong class. In order to mitigate this issue, we tried relabeling the samples and training two separate models. For the first model, we mask the Weak class to ``Other" and train it to predict tags for Strong class and ``other"". Similarly for the second model, the Strong classes are masked to ``other" and a new model is trained this relabelled data. Finally, combine the predictions from the models and the results of this method are included in the Table \ref{tab:final_acc}. It is evident that this relabeling method does not improve the results and thus motivates to follow a novel method of classification using RNN-CNN followed by NER prediction. 

\subsection{BERT Embedding with Feed-Forward Layer}
In recent years, the approaches using pre-trained and fine-tuned models for solving natural language processing tasks have become more popular than ever. Large neural networks were trained on general tasks, such as masked language model or next sentence prediction, and then fine-tuned for other tasks. BERT (Bidirectional Encoder Representations from Transformers)\cite{BERT} is a model developed with that spirit. By using many Transformer encoder layers stacked on top of each other, BERT is known to have a deeper sense of natural language context. Therefore, BERT has achieved state-of-the-art results in many tasks, including questions answering, named entity recognition, etc. 

In this project, we experimented BERT fine-tuning for NER task. Firstly, the input text sentence will be fed into BERT pre-trained model (\textit{``bert-base-cased''}), and the model returns an embedding vector for every token in the input sentence. In the second step, the embedding vector of each token retrieved from BERT will be forwarded into a simple linear neural layer for predicting the NER label. We fixed the BERT parameters and fine-tune the later linear layer on the given corpus.  \textit{Note: due to the limited time of this project and the complexity of BERT model, we do not re-implement the model, but rather utilize the implementations of the Hugging Face\footnote{https://github.com/huggingface/transformers} team for experiments.
}

\subsection{Classification Sentence with RNN-CNN} 
This task aims at predicting the presence of the Weak class (”Art”,  ”Eve”  and  ”Nat”)  given a sentence guided by a binary classification model. In other words, instead of composing a single model to process for all classes, our effort focuses on modeling a couple of models of BI-LSTM-CRF such that each of them trained and optimized individually on two separate subcategories: Weak and Strong class. If the binary model returns a zero value it means that the output of BI-LSTM-CRF built on Strong class will also be the final predictions. Otherwise, we need to vote the decisions of two models through some defined rules, eg. piece-wise maximum probability prediction. The idea behind this strategy is with separate models; we can achieve better performance on the Weak class when the bias on the Strong class is eliminated. Furthermore, we can adapt properly to particular prior knowledge on each category to maximize both likelihoods learned from data and posterior distribution.

The desired binary classification model can be cast as a sentiment analysis task \cite{Raffel}, \cite{Sun}, in which given an input sentence, that comprises a list of words, we would like to evaluate its meaning is “positive" or “negative". In our setting,  “positive" is equivalent to the presence of the Weak class and “negative" means that we can not find any element in this class. The data for the training step can be created by scanning entire original data and then marking each sample is “1” if its ground truth contains one of three classes: ”Art”,  ”Eve”  or  ”Nat”.  Otherwise, the sample is labeled as "0". Unfortunately, we again encounter the second problem when the number of positive samples is $588$, which is just approximately $1.78\%$ of the total negative samples with $32982$ instances. 

We examine three current state-of-the-art methods in sentiment analysis \cite{Barnes}, that involves: RNN-CNN, LSTM Attention, and Self-Attention. The detail on these architectures can be found in-works \cite{Vaswani}, \cite{Yang}, \cite{Xie}. For each method, we implement binary cross entropy (BCE) and saving the best model by tracking the  performance after each epoch on the validation set. The chosen measure for tracking is Accuracy on “1” class since we expect the trained model can predict a precisely Weak class "1" than the “0” class which involves both Strong class and ‘O' type. To represent features for each word, we utilize a glove vector embedding \cite{Glove} with 100 dimensions, that trained on Wikipedia 2014 \footnote{https://en.wikipedia.org/wiki/2014} and Gigaword 5 \footnote{https://catalog.ldc.upenn.edu/LDC2011T07} with over 6 Billion tokens, 400,000 vocabularies. 

A naive way of choosing the training set is to seek the equilibrium between "1" and "0" classes. However, we also want not to ignore too much "0" data; consequently, in the first experiment, we employed both two approaches. Table 1 displays the performance we achieved for one and zero class by using either balanced and full training data. The results indicate that both RNN-CNN, LSTM-Attention and Self-Attention work on "0" class better than "1" class and its tendency to be improved when using full data set.  Even so, in both cases, the best outcomes for the "1" class were still very low with approximately $55\%$.

To overcome this issue, we introduce a modified binary loss function that exploits the whole data for training procedure:  
\begin{equation}
    L_{wce} = -p_{0}t_{0}\log(s_0)  - p_{1}t_{1}\log(s_1)
\end{equation}
where $(t_0, s_0), (t_1, s_1)$ denote for the ground truth and scores for class $0$ and $1$, respectively. $p_0, p_1$ denote for the occurrence probability of class $0$ and $1$ in the training data. In our experiment, the $p_0 \approx 0.02$ and $p_1 \approx 0.98$. Table \ref{tab:table3} presents the performance of three baseline models under this setting. It is really interested when the RNN-CNN improves the accuracy on both $0$ and $1$ class with a large margin compared two remaining methods, holding $75 \%$ on "1" class and $99.6 \%$ on "0" class. This improvement can be reasoning as the effect of setting small weights for class 0 and large weights for class 1. It forces the model to damp loss values when predict correctly "0" class and amplifying error for incorrect assignment on "1" class.  

\subsection{Majoring Voting} 
To make a final prediction, we need to derive three model: one for binary classification described as in section 2.3 and two separate trained models for Strong class (``Geo'', ``Time'', ``Org'', ``Per'' and ``Gpe'')  and the Weak class (``Art'', ``Eve'' and ``Nat''). Denote $P_f$, $P_{str}$, $P_{w}$ $B_w$,  are the output of final prediction, Strong classifier, Weak classifier and binary classification model. Given a sentence $s = [s_1, s_2, ..., s_n]$, the $P_f(s)$ is computed by:
\begin{equation}
    P_f (s)= \left\{\begin{matrix}
P_{str} (s) \ \  \text{if}  \ B_{w}(s) = 0\\ 
P_{w}(s) \odot \ P_{str}(s)\ \ \text{otherwise}
\end{matrix}\right.
\end{equation}
where 
\begin{equation}
(P_{w}(s) \odot \ P_{str}(s))_{i} = \left\{\begin{matrix}
P_{w}(s_{i})\ \ \text{if}  \ \ P_{str}(s_{i}) = \text{O} (\text{other})\\  
P_{str}(s_{i})\ \ \text{if}  \ \ P_{w}(s_{i}) = \text{O} (\text{other})\\ 
P_{m}(s_{i})\ \ \text{otherwise} 
\end{matrix}\right.
\end{equation}
with $ m = \text{argmax}_{j \in {(w, str)}} {[p_w(s_i), p_{str}(s_i)]}$,  $p_w(s_i)$ and $p_{str}(s_i)$ are probability output given by Strong and Weak classifier at $s_i$ position in the $s$ sequence.
 
%-------------------------------------------------------------------------
\section{Result}
Table \ref{tab:final_acc} illustrates two our different methods based on Bi-LSTM-CRF and BERT- Feedforward. We also compared the derived results to four other baselines include original Bi-LSTM-CRF, BERT- Feedforward and Double version of each architecture. In Double version of each architecture, we train one model for weak classes, one model for strong classes and finally merge two labels to form the final prediction using a simple merging schedule. Sentence type classification is not used in this version. In the result table, the bold value denotes for the best value cross all methods. For each entity category, we compute the $F_1$ score and using global $F_1$,  weighted average $F_1$ and macro-average of $F_1$ for all classes by the following formula:

\begin{equation}
    \text{Global\ } F_{1} = \frac{2 * Precision * Recall}{Precision + Recall}
\end{equation}
where
\begin{equation}
    Precision = \frac{TP}{TP +FP}
\end{equation}
\begin{equation}
    Recall = \frac{TP}{TP + FN}
\end{equation}
with TP is true positive, TN is true negative, FP is false positive and FN is false negative computed globally . Macro-Average F1 and Weighted Average F1 are given by:
\begin{equation}
    \text{Macro-Avg} \ F_{1} = \frac{1}{K}\sum_{i=1}^K F_{{1}_i} 
\end{equation}
\begin{equation}
    \text{Weighted Average} \, F_{1} = \frac{\sum_{i=1}^K n_iF1_i}{\sum_{i=1}^K n_i}
\end{equation}
with $K$ is the number of classes, $n_{i}$ is total samples in each class $i$ and $F_1i$ is the F1 score for class $i$.

\begin{table}
\label{tab:table2}
\begin{center}
\scalebox{1.0}{\begin{tabular}{|c|c|c|c|c|}
\hline
\textbf{Method} & \multicolumn{2}{c|}{\textbf{0-Class}} & \multicolumn{2}{c|}{\textbf{1-Class}}\\
\cline{2-5}
 & Balanced & Full  & Balanced  & Full \\
\hline\hline
LSTM - Attention &  83.0 & 98.4  &43.0  &\textbf{55.5} \\
Self-Attention & 85.8 & 98.7  &\textbf{45.3} &50.0\\
\textbf{RNN-CNN }& \textbf{92.7} & \textbf{99.1} &29.7 &53.1\\
\hline
\end{tabular}}

\end{center}
\caption{Accuracy of baseline methods using either full and balanced training data without  weighted loss function.}
\end{table}

\begin{table}
\begin{center}
\begin{tabular}{|c|c|c|}
\hline
\textbf{Method} & \textbf{0-Class} & \textbf{1-Class}\\
\hline\hline
LSTM - Attention & 98.7  &  60.1\\
Self - Attention &  99.1   & 54.7 \\
\textbf{RNN - CNN} & \textbf{99.6} & \textbf{75.0} \\
\hline
\end{tabular}
\end{center}
\caption{Accuracy of baseline methods using full training data with weighted loss function.}
\label{tab:table3}
\end{table}

First of all, we can perceive that our approach based on the BERT-FeedForward version holds the best accuracy on both three separate metrics: global, weighted and macro average compared to other rivals. Especially, we obtain a large margin on the Macro average with 0.67, higher than the second-highest one by 0.03 and the worst method up to 0.12. In other ways, the last result also confirmed the proposed method benefits in improving the performance of the Weak class significantly. For instance, the F1 score for Nat, Art, Eve groups in Bi-LSTM-CRF, increased from 0.26, 0.00, 0.17 to 0.43, 0.10, 0.24 respectively. So on, these values have raised from 0.42, 0.17, 0.30 to 0.42, 0.34, 0.37 for model based on BERT-FF. In summary, these statistical values have shown the effectiveness of our method, not only to help preserve accuracy in classes with many labels but also to significantly increase correctness in classes with just a few labels. 

\begin{table*}[t]

\begin{center}
\scalebox{0.9}{%
\begin{tabular}{|c|c|c|c|c|c|c|}
\toprule
{\textbf{Method}} & \multicolumn{1}{c|}{\textbf{Bi-LSTM-CRF}} & \multicolumn{1}{c|}{\textbf{BERT - FF}} & \multicolumn{1}{c|}{\textbf{Double Bi-LSTM-CRF }} & \multicolumn{1}{c|}{\textbf{Double BERT - FF}} & \multicolumn{1}{c|}{\textbf{Our Method \footnote{Bi-LSTM-CRF}}} & \multicolumn{1}{c|}{\textbf{Our Method} \footnote{BERT - FF}}\\
\midrule
\hline
\textbf{Tim} & 0.85 & \textbf{0.88} & 0.84 & \textbf{0.88} & 0.84 & \textbf{0.88}\\
\textbf{Per} & 0.69 & \textbf{0.78} & 0.67 & \textbf{0.78} &	0.67 & \textbf{0.78} \\
\textbf{Geo} & 0.84 & \textbf{0.88} & 0.84 & 0.87 & 0.84 & 0.87 \\
\textbf{Org} & 0.64 & 0.70 & 0.64 & \textbf{0.72} & 0.64 & \textbf{0.72} \\
\textbf{Gpe} & 0.95 & \textbf{0.96} & 0.95 & \textbf{0.96} & 0.95 & \textbf{0.96} \\
\textbf{Nat} & 0.26 & 0.42 & 0.20 & 0.12 & \textbf{0.43} & 0.42 \\
\textbf{Art} & 0.00 & 0.17 & 0.01 & 0.05 & 0.10 & \textbf{0.34} \\
\textbf{Eve} & 0.17 & 0.30 & 0.00 & 0.00 & 0.24 & \textbf{0.37}\\
\textbf{All classes} & 0.80 & \textbf{0.84} & 0.75 & 0.83 & 0.77 & \textbf{0.84} \\
\textbf{Weighted Average} & 0.79 & \textbf{0.84} & 0.79 & \textbf{0.84} & 0.79 & \textbf{0.84}  \\
\textbf{Macro Average} & 0.55 & 0.64 & 0.52 & 0.55 & 0.59 & \textbf{0.67}\\
\bottomrule
\end{tabular}}
\end{center}
\caption{Comparing our methods to baseline based on Bi-LSTM-CRF and BERT-FeedForward on F1 metric. The bold value indicates for the best result. The first version $\text{Our Method}^{4}$ is modified from Bi-LSTM-CRF and the other based on BERT-FeedForward.} 
\label{tab:final_acc}
\end{table*}

\section{Discussion}
From the above analysis of the results, it can be seen that Bi-LSTM-CRF and BERT architecture have archived promising performance on the task of named entity recognition. Bi-LSTM-CRF accomplished lower F1 score on test set in comparison with BERT architecture, however BERT took much more time for training, fine-tuning and inference due to its high complexity. Therefore, both approaches are useful in different circumstances where the trade-off between running time and accuracy is taken into consideration. 

All methods we have tried suffered from the problem of highly skewed label distribution. From our perspective, there is no method which can perfectly resolve this issue. We attempted in solving the problem by divide and conquer approach, separating Weak classes from Strong classes and training different models for each group. However, the better solution is still to collect more annotated data for the Weak classes. 

During the training process, we realized that there are many wrong annotations in the train/test/dev data. These imperfect annotations might affect the performance of the models, especially in the case of week classes.  Due to limited time of this project, we have not investigated on handling this issue. However, this might be a direction for our future work.   
\section{Conclusion}
In this project, we have proposed a novel procedure to handle sequences labeling problems (NER task) under a highly unbalanced training set. While recent state-of-the-art methods based on Bi-LSTM CRF using embedding features from external resources like BERT just reached a poor performance on categories with fewer training samples, our method can enhance results significantly under such conditions by composing a classification component to locate whether a sentence involves Weak classes. Experiments on test set prove the effectiveness of our method when the score for all Weak classes improved over 50\% compared to the baseline. Directions for the future works include investigating prior knowledge into the classification model and jointly training together two model sequences labeling under specific constraints to make our method be more robust under noise and missing data.

\end{document}